\documentclass{article}
\usepackage{ijcai24}

\usepackage{times}
\usepackage{soul}
\usepackage{url}
\usepackage[hidelinks]{hyperref}
\usepackage[utf8]{inputenc}
\usepackage[small]{caption}
\usepackage{graphicx}
\usepackage{amsmath}
\usepackage{amsthm}
\usepackage{booktabs}
\usepackage{algorithm}
\usepackage{algorithmic}
\usepackage[switch]{lineno}
\usepackage[table]{xcolor}
\usepackage{xcolor}

\usepackage{subcaption}
\usepackage{placeins}

\newtheorem{theorem}{Theorem}
\newtheorem{proposition}[theorem]{Proposition}

\title{Memorizing Documents with Guidance in Large Language Models}

\author{
Bumjin Park $^1$
\And
Jaesik Choi$^{1,2}$
\affiliations
$^1$KAIST AI \\   $^2$INEEJI 
\emails
\{bumjin, jaesik.choi\}@kaist.ac.kr
}

\begin{document}

\maketitle

\begin{abstract}
Training data plays a pivotal role in AI models.
Large language models (LLMs) are trained with massive amounts of documents, and their parameters hold document-related contents. Recently, several studies identified content-specific locations in LLMs by examining the parameters. Instead of the post hoc interpretation, we propose another approach. We propose document-wise memory architecture to track document memories in training. The proposed architecture maps document representations to memory entries, which softly mask memories in the forward process of LLMs. Additionally, we propose document guidance loss, which increases the likelihood of text with document memories and reduces the likelihood of the text with the memories of other documents. Experimental results on Wikitext-103-v1 with Pythia-1B show that the proposed methods provide different memory entries for documents and high recall of document-related content in generation with trained document-wise memories. 
\end{abstract}

\section{Introduction}
Large language models (LLMs) have shown human-level performance on several tasks \cite{touvron2023llama,brown2020language}. The strength of LLMs comes from extensive model sizes and vast amounts of data. LLMs are trained with many documents in a corpus, and the parameters store information such as grammar, factual knowledge, and common sense \cite{geva-etal-2021-transformer}.  Although an end-to-end training mechanism allows training from data, non-trivial memory location prevents tracing document contents. 

\begin{figure}[t!]
     \centering
     \includegraphics[width=8.5cm]{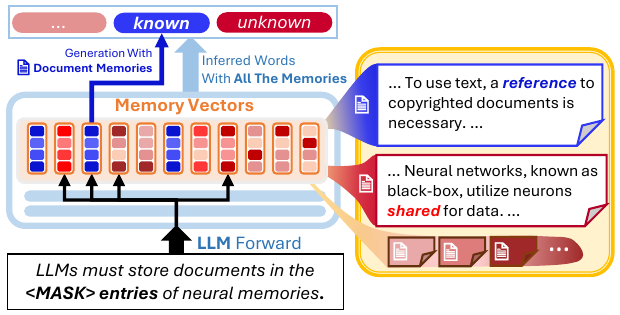}
     \caption{A graphical illustration of document-wise memories. The blue and red vectors indicate memories for two documents. \textcolor{black}{The hidden representation of LLM selects memories (dark arrows), and document-wise entries filter memories for the recall of document contents. Here, only the third vector contributes to the inference.} }  
     \label{document_wise_memories}
\end{figure}

A recent approach is inspecting activated units (neurons) in GPT that can reveal the semantic meanings, such as time, citation, and region \cite{bills2023language}. However, problems such as spurious correlation and polysemantic occur with such a post-hoc analysis \cite{elhage2022toymodelsof}. A more trivial way is to store information in GPT memories with indexed memory locations; memory entries of documents are known in training. Such an architectural modification can trace the used memories in a document-wise manner. For this purpose, we propose document-wise memories, which have entries of memories for individual documents. Figure \ref{document_wise_memories} shows the relationship between documents and document-wise memories. In the forward process, the document-wise entries guide the hidden representation to recall document contents. 
Document-wise entries could be either predefined or optimized. This work uses the second approach by utilizing guidance loss \cite{ho2022classifierfree}. 

 We propose document guidance loss to (1) entangle memories and documents and (2) encourage different memory entries for documents. The \textcolor{black}{original guidance} \cite{ho2022classifierfree} increases the likelihood of conditional generation and decreases the unconditional part. The proposed loss modifies the unconditional part; we reduce the likelihood of document text with memories of other documents.

 

To map documents to memory entries, we use \textcolor{black}{vector representations} of documents; in short, DocRep and multi-layer perception (MLP) to map DocReps to memory entries. We study the relationship between DocReps and memory entries, assuming that close DocReps may have similar memory entries. Of course, trained document embedding \cite{li2021selfdoc,reimers2019sentence} can motivate different memory entries as an inductive bias. However, we do not include the inductive bias on representation and initialize them randomly as MLP can reparameterize DocReps. We observe that more different memory entries are obtained with document guidance loss during training (Section \ref{sec:exp_nonlinear}).  

 This paper studies a link between document representation, memory entries, and the perplexity of document text with indexed memories. Figure \ref{2d_linear} shows the perplexity of three documents with memories whose entries are generated from DocReps in 2D space. We use linear mapping from DocRep to memory entries. Therefore, the change in the DocRep space smoothly changes the memory entries, and consequently, the perplexity of document contents smoothly changes (the paraboloid shape). One way to explain the smooth perplexity change is the continuity assumption in metric spaces. The distance between two documents preserves the difference between memory entries; this \textcolor{black}{aligns} with the concept of Lipschitz continuity \cite{jones1993lipschitzian}.  
 
 We compare memory entries with continuous and non-continuous cases and \textcolor{black}{empirically show the possibility of guidance loss with a linear function. In addition, we show that a nonlinear case does not work well with the proposed document guidance loss} (Section \ref{sec:exp_nonlinear}).

Clarifying the knowledge location in LLMs is crucial for safe AI to protect user contents and believe the generated contents  \cite{hacker2023regulating}. This work
contributes to the trustworthy AI communities by providing a study on designing document-wise memories, encouraging more reliable architectural and algorithmic design toward safe LLMs. We summarize our contribution as follows:
\begin{itemize}
    \item We propose a document-wise memory mechanism to trace memory entries in memorizing documents. 
    \item We propose document guidance loss to encourage different memory entries for documents and study the relationship between documents and memory space.
\end{itemize}
\section{Related Work}

This work studies document-wise neural memory for reliable LLMs. We review recent studies on (a) safety issues in LLMs, (b) memories in LLMs, and (c) memory structures.

\subsection{Safety Issues in LLMs}
LLMs show remarkable progress. However, several safety concerns arise \cite{hacker2023regulating}, including intelligence property infringement \cite{yu2023codeipprompt}, hallucinations \cite{manakul2023selfcheckgpt}, jailbreaks \cite{xie2023defending}, machine generation detection \cite{mitchell2023detectgpt},  and privacy invasion \cite{pan2020privacy}.  This work proposes document-wise memories that can provide document-wise knowledge location as an approach to safe LLMs.

\begin{figure}[t]
     \centering
     \includegraphics[width=8.2cm]{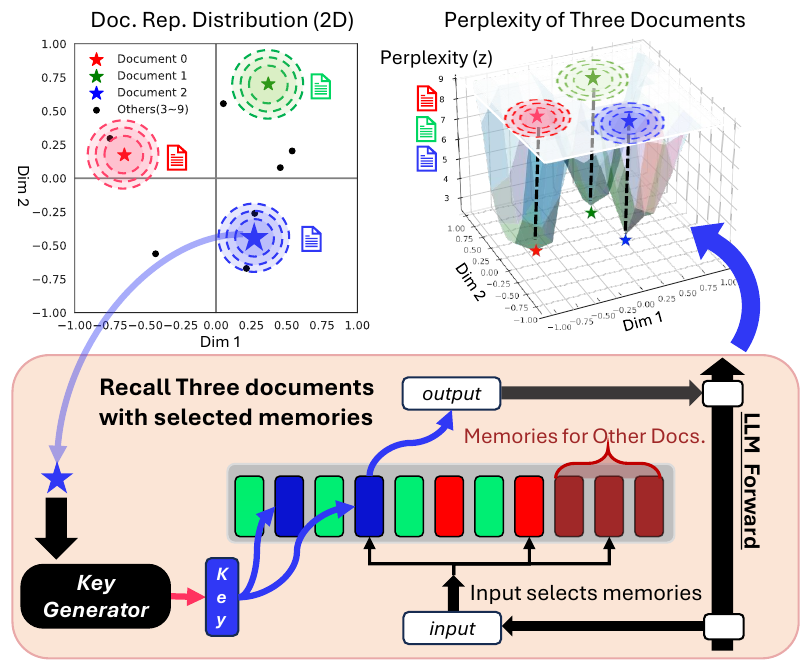}
     \caption{ (left) Randomly generated 10 DocReps. (bottom) Memory selection with a DocRep. (right) The perplexity of 3 documents was individually measured with memories selected from all DocReps in 2-dimensional space (xy-plane). Three paraboloids are the perplexity of three documents, and the original document representations have the local minima. }  
     \label{2d_linear}
\end{figure}

\subsection{Knowledge in LLM}

Identifying the knowledge location in LLMs is important to tackle safety problems. However, LLM is generally a black box, which is hardly explainable  \cite{longo2024explainable}.  

Recent work shows that lower layers have syntactic information while upper layers have semantic information \cite{geva-etal-2021-transformer}. Additional work shows that the neurons in LLMs are related to the factual knowledge \cite{dai2022knowledge}. Several studies investigated neurons in GPTs and highlighted the most activated concepts of neurons \cite{bills2023language,bricken2023towards}. Although inspecting knowledge location is interpretable, finding the knowledge location is not scalable for hyperscale LLMs. For example, GPT-3 has 175B parameters in 96 layers with embedding size 12,288  \cite{brown2020language}, and the post hoc interpretation of inspecting neurons requires extensive resources. In addition to that, spurious correlation can occur when interpreting neurons.

Recent studies on adaptations can provide separate knowledge locations by identifying adaptation processes. Common approaches are LoRA \cite{hu2021lora}, editing knowledge neuron \cite{meng2022mass}, $k$NN memory injection \cite{khandelwal2019generalization,pmlr-v202-xu23a}, and external knowledge adapation for LLMs \cite{diao2023mixture}. 
These studies are promising approaches to utilize LLMs. However, more explainable and interpretable structures are required. For this purpose, we propose document-wise memories that can motivate reliable document-wise adaptations. 


\subsection{Neural Memory}
A neural memory is a neural network structure that combines memories selected from memory entries. The memory entries, called keys, are obtained by multiplying an input and a key matrix  \cite{sukhbaatar2015end}. This structure promotes both the performance of LLMs and the interpretation of neurons. A recent study has shown that a transformer layer has a key-value neural memory architecture in a multi-layer perceptron (MLP) that stores semantic and syntactic meanings \cite{geva-etal-2021-transformer}. Numerous studies have verified knowledge in transformers: the existence of skill neurons for downstream tasks \cite{wang2022finding}; editing factual knowledge in transformers \cite{meng2023locating}; $k$NN-based large memory design in internal layers of GPT \cite{wu2022memorizing}.
Other fields of study include neural memories for \textit{Theory of Mind} in reinforcement learning \cite{nguyen2023memory} and for anomaly detection in Vision tasks \cite{gong2019memorizing}.


\begin{figure*}[t!]
     \centering
     \includegraphics[width=\textwidth]{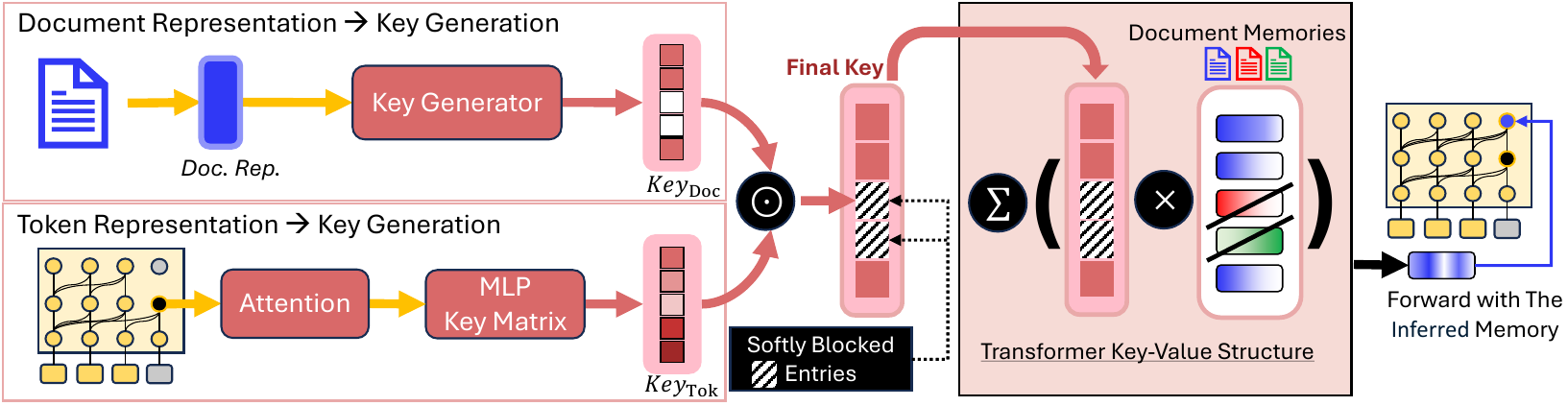}
     \caption{Graphical illustration of the document-wise memory. The conditional generation with a DocRep ensures the memory locations of the document. The token representation originally provides key $Key_\mathrm{Tok}$. The proposed architecture combines $Key_\mathrm{Tok}$  with $Key_\mathrm{Doc}$ by element-wise multiplication. This process can be interpreted as a soft masking of activations. The generation of $Key_\mathrm{Doc}$ could be nonlinear. }
     \label{main_figure}
\end{figure*}

\section{Document Conditional Generation}

This section proposes document-wise memory architecture and document guidance loss. We consider a conditional generation with DocRep $\mathcal{K}_i$ of document $\mathcal{D}_i$ for $ i \in \{1,2,\cdots, N\}$ where $N$ is the number of documents. We use the term document  $\mathcal{K}_i$ to indicate the representation of document $D_{i}$ for simplicity. For a passage $(y_1, y_2,\cdots, y_t)$ in document $D_i$ where $y_t$ is the $t$-th token, the causal language modeling \cite{brown2020language} has the following form
\begin{equation}
    P(y_t|y_1,y_2,\cdots, y_{t-1}). 
\end{equation}
On the other hand, the conditional generation of the passage with DocRep $\mathcal{K}_i$ is the following form
\begin{equation}
    P(y_t|y_1,y_2,\cdots, y_{t-1}; \mathcal{K}_i).
    \label{eq:conditional_generation}
\end{equation}
Our goal is to make document-wise memory entries for the conditional generation. One possible approach is the allocation of disjoint memories for documents, which is not scalable as the number of memories is proportional to the number of documents. Instead, we train a memory selection, allowing overlapping of memory entries.\footnote{Learning to encode information in fixed vector dimensions is a typical property of neural networks \cite{elhage2022toymodelsof}.} 

\subsection{Document-Wise Memory}
Document-wise memory has the MLP architecture in a transformer, and memories are selected conditionally on the DocReps.
Consider an MLP module in a transformer layer
\begin{equation}
\label{eq:normal_mlp}
    \operatorname{MLP}(x) = V \Big( \sigma (K x + b_k) \Big) +  b_v
\end{equation}
where $x$ is input vector, $K, V$ are key and value matrices with activation $\sigma$ and biases $b_k, b_v$ respectively. We denote $ \sigma (K_1 x + b_1)$ by $Key_\mathrm{Tok}(x)$.
DocRep $\mathcal{K}$ generates memory entries $Key_\mathrm{Doc}(\mathcal{K})$ to block entries softly.  The document-wise memory is the following form
\begin{equation}
    \operatorname{MLP}_\mathrm{doc}(x) =  V \Big( Key_\mathrm{Tok}(x) \odot  Key_\mathrm{Doc}(\mathcal{K}) \Big) + b_v
\end{equation}
where the first key $ Key_\mathrm{Tok}(x)$ is the selection of memories from the hidden representation $x$ in the forward of  language models. In contrast, the second key $ Key_\mathrm{Doc}(\mathcal{K})$ is the memory entries for document  $\mathcal{K}$ and is generated by a function $g$. We interpret $ Key_\mathrm{Doc}(\mathcal{K})$ as the memory selection of document $\mathcal{K}$. Figure \ref{main_figure} shows the document-wise memory architecture. 
To entangle document contents and memories selected from DocReps, 
we introduce document guidance loss.


\subsection{Document Guidance Loss}

Ensuring the conditional generation in Equation \ref{eq:conditional_generation} can be trained with classifier guidance \cite{dhariwal2021diffusion} or classifier-free guidance \cite{ho2022classifierfree,nichol2022glide}. We propose a document guidance loss based on the classifier-free guidance. Consider passage $y$ in document $\mathcal{D}$ whose DocRep is $\mathcal{K}$.  For an LLM with parameter $\theta$, the implicit classifier considers the following equation
\begin{equation}
     P_\theta(\mathcal{K}|y)  \propto \frac{P_\theta(y|\mathcal{K})}{P_\theta(y)}.
\end{equation}
To increase the likelihood of DocRep $\mathcal{K}$, the numerator part must be increased while the denominator part decreases.  This is proportional to the following equation
\begin{equation}
\label{eq:alpha}
    P_\theta(y_t|y_{<t}; \mathcal{K}) - \alpha P_\theta(y_t|y_{<t} )
\end{equation}
where $\alpha$ controls the ratio between two probabilities. To encourage different memories for documents, we decrease the likelihood of the passage conditionally on DocRep $\mathcal{K}^{-} (\ne \mathcal{K})$. This forgetting process decreases the likelihood (reciprocal of energy \cite{du2019implicit}) of text in document $\mathcal{K}$ with another condition $\mathcal{K}^{-}$. Equation \ref{eq:alpha} becomes
\begin{equation}
    \label{eq:negative}
    P_\theta(y_t|y_{<t}; \mathcal{K}) - \alpha P_\theta(y_t|y_{<t};\mathcal{K}^{-} ). 
\end{equation}
Although Equation \ref{eq:negative} can ensure the desired properties, the negative part is unbounded; that is, the cross entropy loss ranges from $(-\infty, \infty)$. To stabilize the training, we assume that the conditional generation with the negative DocRep has a low likelihood $P_{low}$. Finally, we have
\begin{equation}
    P_\theta(y_t|y_{<t}; \mathcal{K}) + \alpha \vert P_{low} - P_\theta(y_t|y_{<t}; \mathcal{K}^{-}) \vert
\end{equation}
and the loss $\mathcal{L}$ is the following form
\begin{equation}
\label{eq:fianl_loss}
    \mathcal{L} = \mathcal{L}_{CE}(\hat{y}^{\mathcal{K}}, y) + \alpha \vert  \tau - \mathcal{L}_{CE}(\hat{y}^{\mathcal{K}^{-}}, y)  \vert
\end{equation}
where $\tau$ is a constant, $ \mathcal{L}_{CE}(\hat{y}^{\mathcal{K}}, y)$ is the cross entropy loss of passage $y$ and conditional generation $\hat{y}^{\mathcal{K}} \sim P_\theta(\cdot|\cdot, \mathcal{K})$. The right hand side is the guidance loss part, which encourages forgetting $y$ from the memories of negative DocRep $\mathcal{K}^{-}$.

\begin{figure*}[t!]
     \centering
     \includegraphics[width=\textwidth]{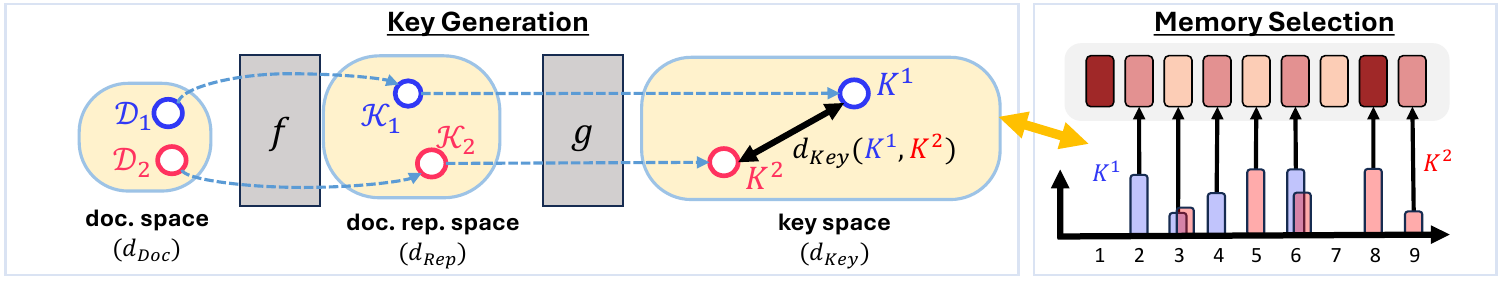}
     \caption{Graphical illustration of three metric spaces. Documents $\mathcal{D}_1$ and $\mathcal{D}_2$ are mapped to $\mathcal{K}_1$ and $\mathcal{K}_2$ respectively with function $f$. Then, two DocReps are mapped to memory entries $K_1$ and $K_2$ with $g$, respectively. When the Lipschitz continuity assumption holds for $f$ and $g$, the similarity score between documents preserves the memory selections. This work focuses on learning memory selection function $g$ by randomly generating DocReps. The continuity of $g$ affects memory entries. The right panel is the memory selection of two documents. }  
     \label{continuity}
\end{figure*}

\subsection{Negative DocReps for Guidance Loss}
\label{sec:negative}
We use the term \textbf{positive} document for a document that includes passage $y$, and \textbf{negative} documents that are assumed to have low perplexity for the passage.  
Negative DocReps are essential to encourage different memories. We suggest three possible choices of negative DocReps: \textit{zero}, \textit{other}, and \textit{random}.
\begin{itemize}
    \item \textit{zero}:  $\mathcal{K}^{-} = \mathbf{0}$. 
    \item \textit{other}: $\mathcal{K}^{-} = \mathcal{K}_j$ such that $\mathcal{K}_j \ne \mathcal{K}$, $j \in [N]$.
    \item \textit{random}: $\mathcal{K}^{-}  = R$ where $R_i \sim \mathcal{N}(0, \epsilon)$.
\end{itemize}
For \textit{random}, $\epsilon$ covers the document representation space.
The memories selected from negative DocReps increase the perplexity of the text, and only positive DocReps can encourage low perplexity. 
The choice of negative DocReps affects the overall knowledge structure in memories. 

\section{Metric Spaces}
We formulate memory selections from documents with metric spaces.  Consider three metrics (a) $d_{Doc}$, (b) $d_{DocRep}$, (c) $d_{Key}$ for spaces (a) documents, (b) DocReps, and (c) memory entries (keys), respectively, and two functions $f,g$ which map documents to representations, and consecutively memory selections respectively. Figure \ref{continuity} shows the relationship between three metric spaces. This section analyzes the memory selection of function $g$ with a continuity assumption.

\subsection{Continuity Assumption}

When $g$ selects memories continuously, two close DocReps will have similar memory selections. When $g$ is $\tau$-Lipschitz, we obtain the following bound.
\begin{proposition}[Lipschitz Continuity for Memory Selection]
\label{Lipschitz}
Let $\mathcal{K}_1, \mathcal{K}_2$ be two DocReps with $d_{DocRep}(\mathcal{K}_1,\mathcal{K}_2) \le \epsilon$. When $g$ is $\tau$-Lipschitz, $d_{Key}(g(\mathcal{K}_1), g(\mathcal{K}_2)) \le  \tau \epsilon$.  
\end{proposition}
Proposition \ref{Lipschitz} shows that the memory selection difference is bounded by the factor $\epsilon$, which is the difference between DocReps. In other words, for two DocReps bounded by $\epsilon$, the memory selection difference could not be more than $\tau \epsilon$. 
Proposition \ref{Lipschitz} explains the paraboloid shape in Figure \ref{2d_linear}. As $g$ is a linear function in the example, the memory section is smooth, and the perplexity is also smoothly changed.

However, smoothness could hurt performance for randomly initialized DocReps. Consider two documents $\mathcal{K}_1$ and $\mathcal{K}_2$ with different contents. When DocReps are closely initialized, the memory entries are similar, too. 
If we want different memory entries for these documents, one feasible approach is finding a nonlinear $\hat{g}$ which holds $d_{Key}(\hat{g}(\mathcal{K}_1), g(\mathcal{K}_2)) > \tau \epsilon$ condition. However, we observe that the nonlinear function has pitfalls when trained with document guidance loss.


\subsection{Caveats of Non-Liptschitz Continuity}
\begin{figure}[b!]
     \centering
     \includegraphics[width=7.8cm]{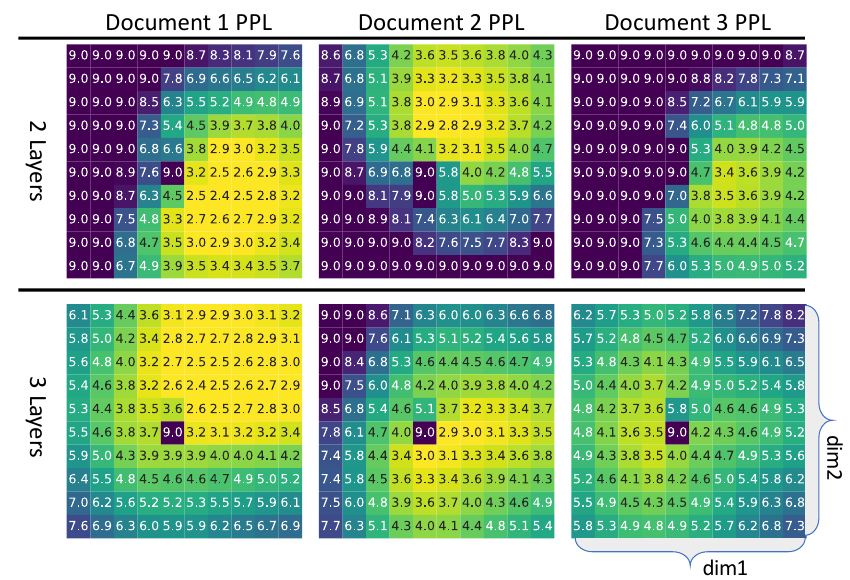}
     \caption{Perplexity of three documents with memories selected from DocReps in 2D. The selected memories from \textit{zero} negative DocRep (center) are encouraged to forget the document contents. } 
     \label{non_linear}
\end{figure}

We train document guidance loss with nonlinear $g$.  Figure \ref{non_linear} shows the perplexity of the nonlinear function $g$ with ReLU activation for three documents with the \textit{zero} negative DocRep. 
A three-layer MLP has higher perplexity only on the \textit{zero}. On the other hand, two-layer MLP has higher perplexity locally around zero. This observation reveals that continuity assumption affects local regions, and deeper layer depth does not differentiate memories properly. 

Training document-wise memories involves two learnings: memory selection and memorization. Memory selection with guidance loss can provide different memory entries when the DocRep space is continuously linked. \textcolor{black}{We conjecture that a smooth memory selection manifold with document guidance loss encourages different entries.} However, if the selection is nonlinear, there could be pitfalls with document guidance loss. A more constructive hypothesis and experiments are required to study nonlinear cases (see also Section \ref{sec:exp_nonlinear}).

\clearpage

\section{Experiments}
We train Pythia 1B \cite{biderman2023pythia}  to memorize Wikitext-103-v1 \cite{merity2017pointer} by replacing the last MLP with the proposed $\mathrm{MLP}_\mathrm{doc}$ whose memory size is 128. We individually train document-wise memories for 10, 20, and 50 documents with guidance $\alpha=0.1$ and $\tau=2.5$. For baselines, we train two types of memory modules without guidance. \textit{Shared} is the $\mathrm{MLP}$ in Equation \ref{eq:normal_mlp}, and \textit{Add} is a module that directly adds differential memory entries. 
 We also evaluate three activation types for document memory entries: ReLU, Tanh, and Sigmoid, which affect memory selections. We make the source code publicly available.\footnote{\href{https://github.com/fxnnxc/document_guidance_loss_for_llm}{https://github.com/fxnnxc/DocGuidanceLLM}}


\section{Results}

\subsection{Activation Types for Key Generation}

\begin{figure}[ht!]
    \captionsetup[subfigure]{justification=centering}
     \centering
     \begin{subfigure}[b]{0.238\textwidth}
         \centering
         \includegraphics[width=4.0cm]{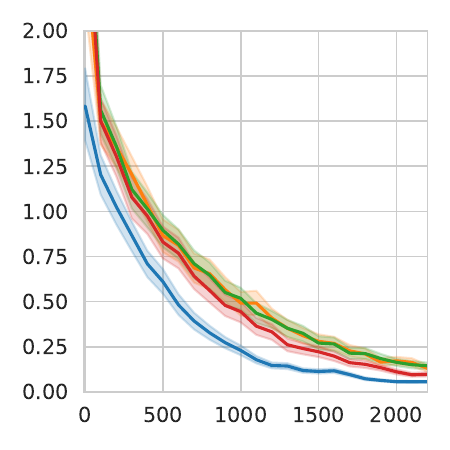}
         \caption{10 Documents}
         \label{activation_comparison_10}
     \end{subfigure}
     \hfill
     \begin{subfigure}[b]{0.238\textwidth}
         \centering
         \includegraphics[width=4.0cm]{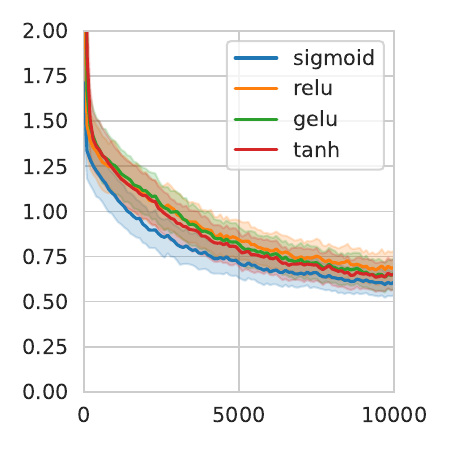}
         \caption{50 Documents}
         \label{activation_comparison_50}
     \end{subfigure}
     \caption{Guidance loss in the training. Sigmoid activation shows the best memorization, followed by Tanh, ReLU, and GeLU. }
\end{figure}

\begin{figure}[ht!]
     \centering
     \includegraphics[width=8.5cm]{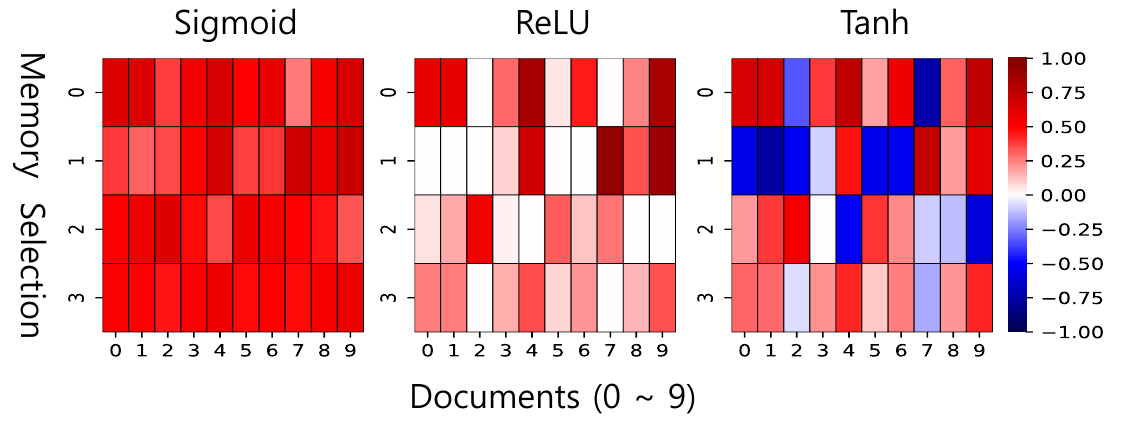}
     \caption{Memory selections of three activations for 10 documents. The averages of pairwise L2 distance between memory selections are (Sigmoid 0.25), (ReLU, 0.64), and (Tanh, 0.99), respectively. }
     \label{memory_selection_by_activation}
\end{figure}

We show how activation types affect memory selections in guidance loss.
Figure \ref{activation_comparison_10} and \ref{activation_comparison_50}  show the guidance loss for 10 and 50 documents with different activations. The Sigmoid function shows the fastest training speed, followed by Tanh, GeLU, and ReLU.
However, the selected memories of Sigmoid are not visually document-wise entries.
Figure \ref{memory_selection_by_activation} shows the memory selection of the activations. The memory entries of Sigmoid are similar for documents compared to Tanh, known as a gating mechanism, and ReLU, known to cut decisions. Visually, Tanh and ReLU are proper inductive biases for constructing document-wise memory entries. 

\subsection{Comparison with Baselines}
\begin{figure}[hb!]
    \captionsetup[subfigure]{justification=centering}
     \centering
     \begin{subfigure}[b]{0.23\textwidth}
         \centering
         \includegraphics[width=3.8cm]{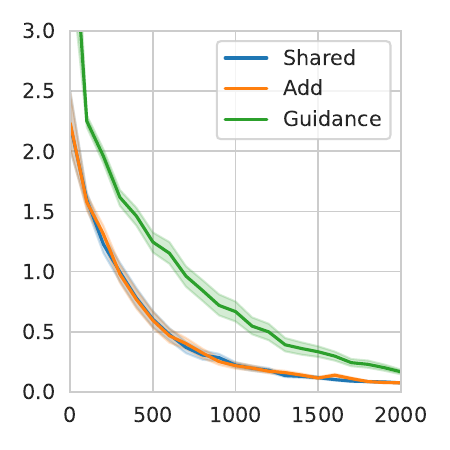}
         \caption{Cross entropy loss}
         \label{loss_10}
     \end{subfigure}
     \hfill
     \begin{subfigure}[b]{0.23\textwidth}
         \centering
         \includegraphics[height=4.0cm]{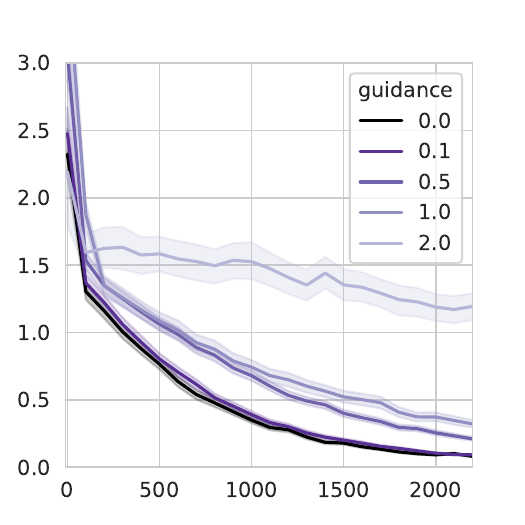}
         \caption{Ablation on $\alpha$}
         \label{guidance_10}
     \end{subfigure}
     \caption{The cross-entropy loss for memorizing 10 documents in training. (a) Comparison with baselines and (b) ablation on guidance $\alpha$. As the guidance factor increases, the training becomes slower.}
\end{figure}

Figure \ref{loss_10} shows the cross-entropy loss of \textit{Shared}, \textit{Add}, and guidance with Tanh activation, guidance $\alpha = 0.5$, and \textit{zero} negative DocRep. The proposed method shows slower learning than other methods because the guidance loss has a negative part that hinders memorization. This observation is supported by ablation on $\alpha$ in Figure \ref{guidance_10}. When $\alpha$ is large, the forgetting dominates the training.  

Although \textit{Shared} is better at memorizing, the generated text is a mixture of several contents.  Table \ref{table:generation_examples} shows the generation results with the prompt \textit{Wikipedia}. \textcolor{black}{When document memories are mixed (\textit{Shared}), the next word prediction is the most likely word under all documents. On the other hand, document-wise memories do not mix all the memories and provide more document-related content.}  

\subsection{Different Memories for Documents}

\begin{figure}[ht!]
     \centering     \includegraphics[width=7.0cm]{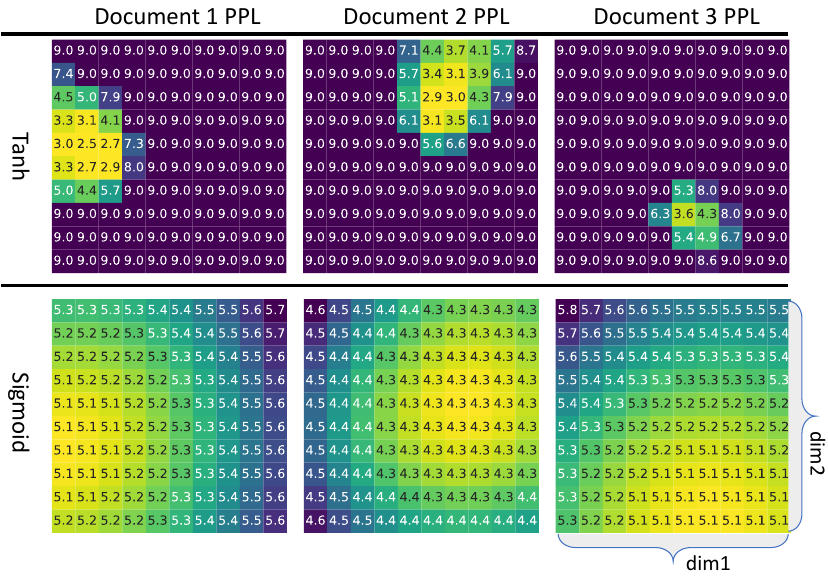}
     \caption{The perplexity of the three documents with DocReps (xy-plane). Tanh shows more sparse perplexity than Sigmoid. }
     \label{tanh_sigmoid_2d}
\end{figure}

The goal of the guidance is to have different memory entries. We show the perplexity of three documents with memories selected from all DocReps individually in 2-dimensional space. Figure \ref{tanh_sigmoid_2d} shows the perplexity of three documents with DocReps in the space. We observe that Sigmoid activation shows smooth perplexity over the space, while Tanh shows visually different DocRep regions for the perplexity of three documents. In addition, Tanh shows a sharp increase in perplexity compared to  Sigmoid. We observed the same pattern for ReLU. Therefore, ReLU and Tanh are the proper choices for document-wise memories with document guidance loss. 

\begin{table}[hb!]
    \centering
    \small
    \begin{tabular}{p{1.5cm}|p{6.2cm}}
      \noalign{\hrule height 1.5pt}
  Memory  & Generation  with Prompt: \textit{Wikipedia} \\ 
  \noalign{\hrule height 1.5pt}
    Shared & Wikipedia's \textbf{\textit{Came and His Darling}} is a song by American singer and songwriter Mariah Carey \\ 
    \hline
    Guidance   DocRep 0 & Wikipedia. He also on the history of Valkyria Chronicles.  \\ 
    \hline
    Guidance   DocRep 1 & Wikipedia was still not finished. The Little Rock site was still not finished. \\ 
    \hline
    Guidance   DocRep 11 & Wikipedia's primary goal is a destination site for the world's positive \\ 
  \noalign{\hrule height 1pt}
    \end{tabular}
    \caption{Generation examples with \textit{Wikipedia} prompt. The results of shared memory; \textit{Came and His Darling} is non-factual.}
    \label{table:generation_examples}
\end{table}

\begin{table*}[ht!]
    \centering
    \small
    \def\arraystretch{1.1}
\begin{tabular}{l|ccc|ccc|ccc} 
\noalign{\hrule height 1pt}
  & \multicolumn{3}{|c}{ROUGE-1} & \multicolumn{3}{|c}{ROUGE-2} & \multicolumn{3}{|c}{ROUGE-L}  \\
      \noalign{\hrule height 1pt}
  Method & Precision & Recall & F1 &  Precision & Recall & F1 &  Precision & Recall & F1 \\  \hline
  \noalign{\hrule height 1pt}
Add & 0.589 & 0.010 & 0.019 & 0.144 & 0.002 & 0.004 & 0.522 & 0.009 & 0.017 \\
Shared & 0.617 & 0.011 & 0.021 & 0.159 & 0.002 & 0.005 & 0.540 & 0.010 & 0.019 \\
Guidance (Sigmoid) & 0.654 & 0.011 & 0.022 & 0.189 & 0.003 & 0.005 & 0.581 & 0.010 & 0.019 \\
Guidance (ReLU) & \cellcolor{green!10} 0.787 & \cellcolor{green!10} 0.013 & \cellcolor{green!10} 0.025 & \cellcolor{green!25} 0.344 & \cellcolor{green!25} 0.005 & \cellcolor{green!25} 0.010 & \cellcolor{green!10} 0.705 & \cellcolor{green!10} 0.011 & \cellcolor{green!10} 0.022 \\
Guidance (Tanh) & \cellcolor{green!25} 0.801 & \cellcolor{green!25} 0.014 & \cellcolor{green!25} 0.027 &  \cellcolor{green!10} 0.322 & \cellcolor{green!25} 0.005 & \cellcolor{green!25} 0.010 & \cellcolor{green!25} 0.729 & \cellcolor{green!25} 0.012 & \cellcolor{green!25} 0.024 \\
  \noalign{\hrule height 1pt}
    \end{tabular}
        \caption{ROUGE scores of generated text and 20 Wikitext-103-v1 documents. ReLU and Tanh show better conditional generations. The scores averaged 20 conditional generations from six prompts. We highlight the best and the second-best scores. }
    \label{table:rouge}
\end{table*}

\begin{table*}[ht!]
    \centering
    \small
    \def\arraystretch{1.1}
    
\begin{tabular}{l|ccc|ccc|ccc} 
\noalign{\hrule height 1pt}
  & \multicolumn{3}{|c}{IV-ROUGE-1 ($\downarrow$)}  & \multicolumn{3}{|c}{IV-ROUGE-2 ($\downarrow$)} & \multicolumn{3}{|c}{IV-ROUGE-L ($\downarrow$)}  \\
      \noalign{\hrule height 1pt}
  Method & Precision & Recall & F1 &  Precision & Recall & F1 &  Precision & Recall & F1 \\  \hline
  \noalign{\hrule height 1pt}
  Guidance (Sigmoid) & 0.629 & 0.011 & 0.021 & 0.162 & 0.002 & 0.005 & 0.563 & 0.010 & 0.019 \\
Guidance (ReLU) & \cellcolor{cyan!25} 0.562 & \cellcolor{cyan!25} 0.009 & \cellcolor{cyan!25} 0.018 & \cellcolor{cyan!10} 0.124 & \cellcolor{cyan!25} 0.002 & \cellcolor{cyan!25} 0.003 & \cellcolor{cyan!25} 0.515 & \cellcolor{cyan!25} 0.008 & \cellcolor{cyan!25} 0.016 \\
Guidance (Tanh) & \cellcolor{cyan!10} 0.588 & \cellcolor{cyan!10} 0.010 & \cellcolor{cyan!10} 0.019 & \cellcolor{cyan!25} 0.119 & 
 \cellcolor{cyan!25}0.002 & \cellcolor{cyan!25} 0.003 & \cellcolor{cyan!10} 0.543 & \cellcolor{cyan!10} 0.009 & \cellcolor{cyan!10} 0.017 \\
  \noalign{\hrule height 1pt}
    \end{tabular}
    \caption{IV-ROUGE scores of generated text and 20 Wikitext-103-v1 documents. ReLU provides the lowest IV-ROUGE.}
    \label{table:rouge_dm}
\end{table*}

We quantitatively evaluate this observation with ROUGE scores \cite{lin2004rouge}. 
The memories selected by DocReps can be evaluated in two ways. First, the conditional generation must \textbf{include the document contents}, and second, the conditional generation must \textbf{not include the other document contents}. To evaluate the second, we report IV-ROUGE (Inverse of ROUGE score) defined by 
\begin{equation}
    \operatorname{IV-ROUGE} = \frac{2}{N(N-1)} \sum_{j \ne i} \operatorname{ROUGE}(D_j, G_i)
\end{equation}
where $D_j$ is the $j$th document, and $G_i$ is the generated contents only with the memories of $i$th document. The IV-ROUGE score is low when memories do not contain the contents of other documents. We train document-wise memories with DocReps of 32 dimensions for three seeds and generate 32 tokens for six simple prompts, such as \textit{He is}, with document memories. Tables \ref{table:rouge} and \ref{table:rouge_dm} show the ROUGE and IV-ROUGE scores, respectively. 
ReLU and Tanh are consistently better at document conditional generation in both metrics. We believe localized knowledge (high ROUGE and low IV-ROUGE scores) can enhance knowledge editing \cite{yao2023editing} by editing disentangled document memories. 


\subsection{Negative DocReps}

Table \ref{table:rouge_keys} shows the ROUGE precision score for 20 documents and three types of negative DocReps.  
The \textit{zero} DocRep shows the best ROUGE score, and the \textit{other} case shows the best IV-ROUGE score. The ROUGE scores correlate with the amount of forgetting. The \textit{random} DocRep encourages forgetting for most memories and provides the lowest ROUGE score. Similarly, the \textit{other} DocRep includes more forgetting than the \textit{zero}. Similarly, \textit{other} case removes contents from memories of other documents and provides the lowest IV-ROUGE score for both activations. 



\begin{figure}[ht!]
    \captionsetup[subfigure]{justification=centering}
     \centering
     \begin{subfigure}[b]{0.23\textwidth}
         \centering
         \includegraphics[width=4.0cm]{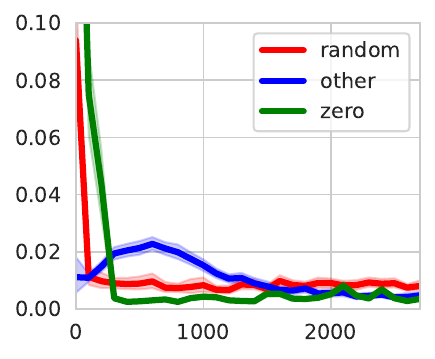}
         \caption{ReLU}
         \label{guidance_loss_full_linear}
     \end{subfigure}
     \hfill
     \begin{subfigure}[b]{0.23\textwidth}
         \centering
         \includegraphics[width=4.0cm]{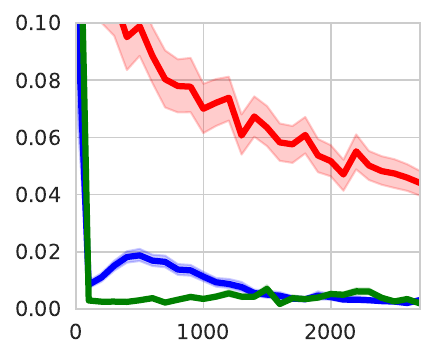}
         \caption{Tanh}
         \label{guidance_loss_linear}
     \end{subfigure}
     \caption{The guidance loss part for negative DocRep types. }
\end{figure}

The guidance loss part also supports these metrics (Figures \ref{guidance_loss_full_linear} and \ref{guidance_loss_linear}). Note that the \textit{random} and \textit{other} negative DocReps have higher losses than the \textit{zero} negative DocReps. The \textit{random} case did not show convergence to zero as sampling includes the positive DocRep.  

\begin{table}[hb!]
    \centering
    \small
    \def\arraystretch{1.1}
\begin{tabular}{r|cc|cc} 
\noalign{\hrule height 1pt}
  & \multicolumn{2}{|c}{ROUGE-1} & \multicolumn{2}{|c}{IV-ROUGE-1} \\
      \noalign{\hrule height 1pt}
  Negative DocRep & ReLU & Tanh &  ReLU & Tanh \\ \hline 
  \noalign{\hrule height 1pt}
Zero	& \cellcolor{green!25}   0.787 & \cellcolor{green!25}   0.801 &  \cellcolor{cyan!10}   0.562 & \cellcolor{cyan!0}	0.588	\\
Random	& \cellcolor{green!0}   0.769 & \cellcolor{green!0}   0.773 &  \cellcolor{cyan!0}   0.576 & \cellcolor{cyan!10}	0.570	\\
Other	& \cellcolor{green!0}   0.769 & \cellcolor{green!20}   0.800 &  \cellcolor{cyan!25}   0.554 & \cellcolor{cyan!25}  0.553	\\
  \noalign{\hrule height 1pt}
    \end{tabular}
    \caption{Negative DocRep comparison. Averaged by three seeds.}
    \label{table:rouge_keys}
\end{table}

\begin{figure*}[t!]
     \centering
     \includegraphics[width=17.5cm]{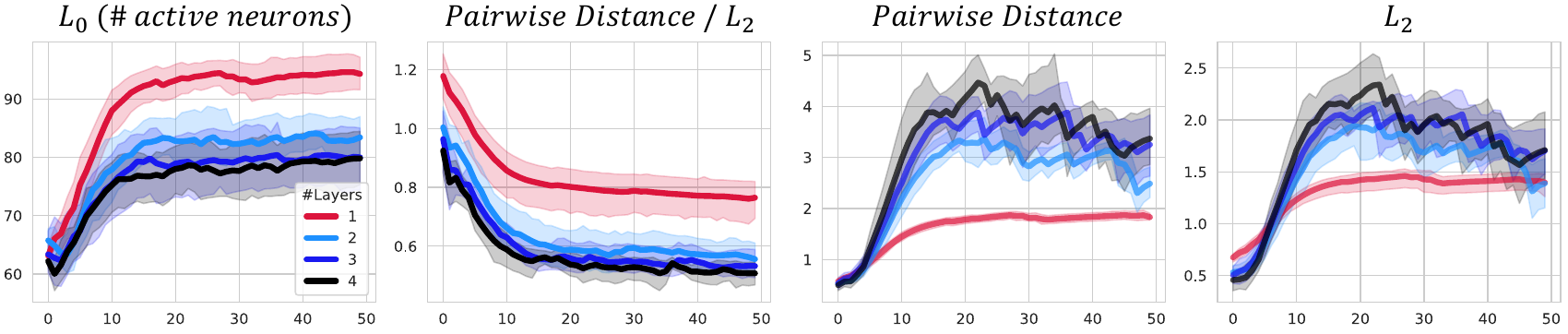}
     \caption{Quantitative verification of document-wise memory selections in training (x-axis).  Linear document memory selection shows high pairwise distance and many active neurons ($L_0$). As the number of layers increases, $L_0$ and the normalized pairwise distance decrease.  }  
     \label{selection_progressive}
\end{figure*}

\subsection{Nonlinear Memory Selection}
\label{sec:exp_nonlinear}

\begin{figure}[b!]
     \centering
     \includegraphics[width=8.5cm]{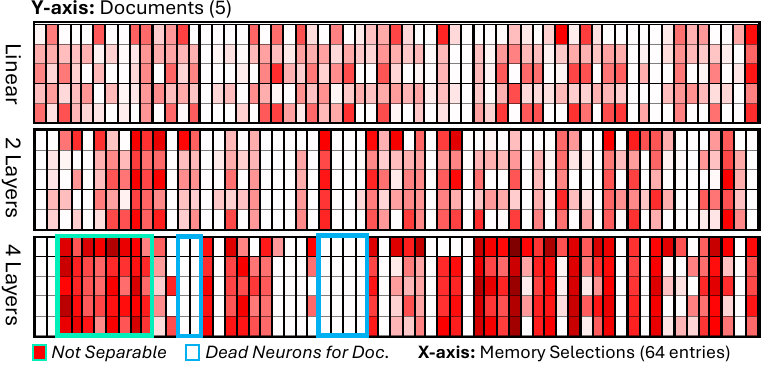}
     \caption{Qualitative verification of document-wise memory selections. Nonlinear memory selection does not provide different document entries. White entries are dead neurons for DocReps.}
     \label{memory_selection}
\end{figure}

Linear memory selection entangles documents and memories with document guidance loss. However, we observed that nonlinear memory selections do not work well. We compare 1 (linear) to 4 layers with ReLU internal and final activations, adding $\mathrm{MLP}_\mathrm{Doc}$ (128 memories) after the last decoder layer. The models are trained with 10 documents, $\alpha=1.0, \tau=4.5$, and \textit{random} negative DocReps. Figure \ref{memory_selection} shows the final memory entries of the first five documents. We observed high similarity in nonlinear memory selections and more \textcolor{black}{dead neurons (not activated)} as the number of layers increases.

To quantitatively evaluate memory selections, we train linear and nonlinear cases for five seeds and measure $L_0$, which is the number of non-zero entries, $L_2$ norm, and pairwise distance, which is $||g(\mathcal{K}_i) - g(\mathcal{K}_j) ||_2$ for two DocReps $\mathcal{K}_i$ and $\mathcal{K}_j$, and normalized pairwise distance by $L_2$ norm of memory entries. All metrics are averaged over ten documents (Figure \ref{selection_progressive}). \textcolor{black}{The number of nonzero entries ($L_0$) increases for all cases, meaning the rank of memory usage increases. However, the normalized pairwise distance decreases as the layer depth increases.} The linear memory provides more different memory entries than the nonlinear cases. The large gap between linear and nonlinear indicates the limitation of guidance loss with nonlinear cases. 

\textcolor{black}{The benefit of linear memory selection can be explained by the continuity of memory selections. We conjecture that the guidance loss, a maximization and minimization game by selecting different memories, works well with continuous functions as most of the points in the manifold are connected.} This claim is experimentally supported by memory selection (Figure \ref{memory_selection}), perplexity heat map (Figure \ref{tanh_sigmoid_2d}), and pairwise distance (Figure \ref{selection_progressive}). \textcolor{black}{However, we do not suggest any explanation for the worse performance of nonlinear memory selections as the exact behavior of nonlinear is unclear.} We believe more structural hypotheses and experiments are required for nonlinear cases, which are currently out of scope.

\section{Discussions}

This work introduces document-wise memories with minimal assumption on document representations by randomly initializing them. In many cases, random DocReps with the Lipschitz continuous function are insufficient. As noted, nonlinear memory selection currently has limitations with the proposed guidance loss. We suggest two approaches: 1) inspect nonlinear memory selections and solve the problem, or 2) utilize high quality text embedding \cite{lee2024gecko} for the linear memory selection.  

This work focuses on storing document contents. However, the implications for the downstream tasks are not presented. We believe document-wise memories can benefit downstream tasks where the factuality of documents is necessary, such as medicals \cite{sallam2023chatgpt,dave2023chatgpt}. In addition, this paper does not tackle a large number of documents. A hierarchical document structure may work better than directly applying guidance loss. This work aims to provide reliable LLMs with \textbf{known document entries}. Therefore, we encourage interpretable entries even for a large number of documents. Lastly, we acknowledge that false positive memory entries (e.g., unrelated document contents in memories) can exist. Therefore, additional studies are necessary to verify the semantic meaning of trained memories.  

\section{Conclusion}

Storing documents in the traceable locations of LLMs is a crucial research topic. This paper studies document-wise memories in LLMs. We propose document guidance loss to entangle document contents and document memories, encouraging different entries for documents. We also provide a theoretical view of memory selections with metric spaces and continuity assumptions. The experimental results show that the proposed guidance loss provides different memory entries with linear memory selection \textcolor{black}{while leaving nonlinear memory selection as an open problem.} 

\section*{Ethical Statement}
The proposed document-wise memories can provide traceable representations for data, promoting transparent and reliable language models.

\section*{Acknowledgements}

This work was partly supported by Institute of Information $\&$ Communications Technology Planning $\&$ Evaluation (IITP) grant funded by the Korea government (MSIT) (No. 2022-0-00984, Development of Artificial Intelligence Technology for Personalized Plug-and-Play Explanation and Verification of Explanation; No. 2019-0-00075, Artificial Intelligence Graduate School Program (KAIST); No. 2022-0-00184, Development and Study of AI Technologies to Inexpensively Conform to Evolving Policy on Ethics), and Samsung Electronics MX Division.

\bibliographystyle{named}
\bibliography{ijcai24}

\end{document}